\documentclass{article}

\usepackage[preprint]{neurips_2026}

\usepackage[utf8]{inputenc} 
\usepackage[T1]{fontenc}    
\usepackage{hyperref}       
\usepackage{url}            
\usepackage{booktabs}       
\usepackage{amsfonts}       
\usepackage{nicefrac}       
\usepackage{microtype}      
\usepackage{xcolor}         
\usepackage{svg}
\usepackage{amsmath}
\usepackage{amsthm}
\usepackage{amssymb}

\usepackage{makecell}
\usepackage{multirow}

\usepackage[noend]{algpseudocode}
\usepackage{algorithm}
\algrenewcommand\algorithmicindent{1.0em}%
\newcommand{\alglinenumberstyle}[1]{{\tiny\color{black!50}#1}}
\algrenewcommand\alglinenumber[1]{\alglinenumberstyle{#1.}\hspace{-2pt}}

\newcommand{\rightcomment}[1]{{\color{gray} \(\triangleright\){\footnotesize\textit{#1}}}}
\algrenewcommand{\algorithmiccomment}[1]{\hfill \rightcomment{#1}}
\algnewcommand{\LineComment}[1]{\State \rightcomment{#1}}
\algnewcommand{\LinesComment}[1]{\State \rightcomment{\parbox[t]{\linewidth-\leftmargin-\widthof{\(\triangleright\) }}{#1}}}

\usepackage{cleveref}
\crefname{figure}{Fig.}{Figs.}
\crefname{table}{Table}{Tables}
\crefname{appendix}{App.}{Apps.}
\crefname{section}{\S}{\S\S}
\crefformat{section}{\S#2#1#3} 
\crefname{equation}{Eq.}{Eqs.}
\crefname{algorithm}{Alg.}{Algs.}
\crefname{algocf}{Alg.}{Algs.}
\crefname{defin}{Def.}{Defs.}
\crefname{theorem}{Thm.}{Thms.}
\crefname{lemma}{Lemma}{Lemmas}

\usepackage[mathscr]{euscript}

\usepackage{tikz}

\usepackage[textsize=tiny,textwidth=1.3in]{todonotes}
\usepackage[most]{tcolorbox}
\newtcolorbox{promptbox}[1][]{
  colback=gray!5,
  colframe=gray!50,
  fonttitle=\bfseries,
  title=Prompt,
  enhanced,
  attach boxed title to top left={yshift=-2mm, xshift=2mm},
  boxed title style={colback=gray!80},
  #1
}

\makeatletter
\newcommand*{\rom}[1]{\expandafter\@slowromancap\romannumeral #1@}
\makeatother

\makeatletter
\newcommand{\oset}[3][0.23ex]{%
  \mathrel{\mathop{#3}\limits^{
    \vbox to#1{\kern-2\ex@
    \hbox{$\scriptstyle#2$}\vss}}}}
\makeatother

\crefformat{equation}{\eqA Eq.~\eqB #2#1#3)}
\newcommand{\eqA}{}
\newcommand{\eqB}{(}

\newcommand{\citeposs}[1]{\citeauthor{#1}'s (\citeyear{#1})}

\definecolor{ETHBlue}{RGB}{33,92,175}   
\definecolor{ETHGreen}{RGB}{98,115,19}      
\definecolor{ETHPurpleDark}{RGB}{140,10,89} 
\definecolor{ETHPurple}{RGB}{163,7,116} 
\definecolor{ETHGray}{RGB}{111,111,111} 
\definecolor{ETHRed}{RGB}{183,53,45}    
\definecolor{ETHPetrol}{RGB}{0,120,148} 
\definecolor{ETHBronze}{RGB}{142,103,19}    
\definecolor{ETHOrange}{RGB}{230, 100, 50}
\definecolor{functionorange}{RGB}{200, 85, 45}
\colorlet{MacroColor}{ETHGreen}
\definecolor{darkgreen}{rgb}{0.0,0.5,0.0}
\definecolor{darkblue}{rgb}{0.0,0.0,0.5}
\definecolor{aggblue}{rgb}{0,0,.65}
\definecolor{vargreen}{rgb}{0,.50,.18}

\newcommand{\mymacro}[1]{#1}

\newcommand{\defn}[1]{\textbf{#1}}

\newcommand{\defeq}[0]{\mymacro{\oset[0.4ex]{\text{\tiny def}}{=}}}

\newcommand{\lang}{\mymacro{L}}
\newcommand{\alphabet}{{\mymacro{\Delta}}}
\newcommand{\signature}{{\mymacro{\Sigma}}}
\newcommand{\premise}{\mymacro{b}}

\newcommand{\goal}{\mymacro{\conclusion'_G}}
\newcommand{\conclusion}{\mymacro{c}}
\newcommand{\genitem}[1]{\mymacro{#1}}

\newcommand{\predicate}[1]{\mymacro{{\color{ETHPurple}\mathrm{#1}}}}
\newcommand{\predicatep}{\mymacro{\predicate{p}}}

\newcommand{\atom}{\mymacro{ t }}
\newcommand{\const}[1]{\mymacro{{\color{darkblue} \textit{\texttt{#1}}}}}

\newcommand{\var}[1]{\mymacro{{\color{vargreen} \textit{\texttt{#1}}}}}
\newcommand{\varx}{\mymacro{\var{X}}}
\newcommand{\vary}{\mymacro{\var{Y}}}
\newcommand{\varz}{\mymacro{\var{Z}}}

\newcommand{\interpretation}{\mymacro{I}}

\newcommand{\constants}{{\mymacro{\signature_{\const{x}}}}}

\newcommand{\variables}{{\mymacro{\signature_{\varx}}}}

\newcommand{\axioms}{\mymacro{A}}

\newcommand{\program}{\mymacro{\mathcal{P}}}
\newcommand{\rules}{\mymacro{\mathcal{R}}}

\newcommand{\worldmodel}{\mymacro{M}}
\newcommand{\generator}[1]{\mymacro{\nu_{#1}}}

\newcommand{\subst}{\mymacro{\theta}}
\newcommand{\substs}{\mymacro{\Theta}}

\newcommand{\herbrandbase}{\mymacro{H}}

\usepackage[group-separator={,},group-minimum-digits={3}]{siunitx}
\newcommand{\proofg}{\mymacro{\pi}}

\newcommand{\vertex}{\mymacro{v}}
\newcommand{\goalvertex}{\mymacro{\vertex_G}}
\newcommand{\source}{\mymacro{S}}

\newcommand{\vertices}{\mymacro{V}}
\newcommand{\edges}{\mymacro{E}}
\newcommand{\edge}{\mymacro{e}}

\newcommand{\cost}{\mymacro{g}}
\newcommand{\weight}{\mymacro{w}}
\newcommand{\goalweight}{\mymacro{\weight^*_G}}

\newcommand{\heuristic}{\mymacro{h}}
\newcommand{\evaluation}{\mymacro{f}}

\newcommand{\outside}{\mymacro{Z}}
\newcommand{\arule}{\mymacro{r}}
\newcommand{\trace}{\mymacro{\boldsymbol{\delta}}}

\newcommand{\popset}{\mymacro{\mathcal{V}_{\text{pops}}}}
\newcommand{\homomorph}{\mymacro{\Phi}}
\newcommand{\homomorphl}{\mymacro{\Phi_{\text{L}}}}

\newcommand{\reward}{\mymacro{r}}

\newcommand{\score}{\mymacro{x}}
\newcommand{\scale}{\mymacro{\beta}}
\newcommand{\location}{\mymacro{\alpha}}

\newcommand{\agenda}{\mymacro{Q}}
\newcommand{\chart}{\mymacro{\mathcal{C}}}

\makeatletter

\newcommand*\wthelper[2]{%
        \hbox{\dimen@\accentfontxheight#1%
                \accentfontxheight#11.2\dimen@
                $\m@th#1\widetilde{#2}$%
                \accentfontxheight#1\dimen@
        }%
}

\newcommand*\accentfontxheight[1]{%
        \fontdimen5\ifx#1\displaystyle
                \textfont
        \else\ifx#1\textstyle
                \textfont
        \else\ifx#1\scriptstyle
                \scriptfont
        \else
                \scriptscriptfont
        \fi\fi\fi3
}
\makeatother

\newcommand{\reals}{\mymacro{ \mathbb R}}
\newcommand{\realspos}{\mymacro{ \mathbb R_{\geq 0}}}
\newcommand{\realskpos}{\mymacro{ \mathbb R^K_{\geq 0}}}

\newcommand{\lm}{\mymacro{p}}

\newcommand{\kleene}[1]{\mymacro{#1^{\!*}}}
\newcommand{\lmalphabet}{\mymacro{\Gamma}}

  {\gdef\scalefactor{#1}\begin{center}\proofSkipAmount \leavevmode}%
  {\scalebox{\scalefactor}{\DisplayProof}\proofSkipAmount \end{center} }

\newcommand{\mydots}{%
  \ifmmode
    .\mkern-1mu.\mkern-1mu.%
  \else
    .\kern-0.1em.\kern-0.1em.%
  \fi
}
\newcommand{\mycdots}{%
  \ifmmode
    \cdot\mkern-1mu\cdot\mkern-1mu\cdot%
  \else
    \cdot\kern-0.1em\cdot\kern-0.1em\cdot%
  \fi
}

\renewcommand{\ldots}{\mydots}
\renewcommand{\dots}{\mydots}
\renewcommand{\cdots}{\mycdots}

\newcommand{\fixpoint}{\mymacro{\mathrm{T}_{\program}}}
\newcommand{\fixpointn}{\mymacro{\mathrm{T}^n_{\program}}}
\newcommand{\fixpointzero}{\mymacro{\mathrm{T}^0_{\program}}}
\newcommand{\fixpointkleene}{\mymacro{\mathrm{T}^*_{\program}}}

\newcommand{\hyperhead}{\mymacro{{\vertex}_c}}
\newcommand{\hyperheadj}[1]{\mymacro{{\vertex}_{c_{#1}}}}

\newcommand{\labeler}{\mymacro{\lambda}}
\newcommand{\premn}[1]{\mymacro{\vertex_{\premise_{#1}}}}

\definecolor{goaltheorem}{HTML}{e3c25b} 
\definecolor{shortestproof}{HTML}{7990d9} 
\definecolor{irrelevanttheorems}{HTML}{d27b77} 

\title{
Learning to Reason Efficiently with A* Post-Training
}

\author{%
  Andreas Opedal$^{\alpha,\beta}$
  \quad Francesco Ignazio Re$^{\alpha}$  
  \quad \textbf{Abulhair Saparov}$^{\gamma}$ \\
  \textbf{Mrinmaya Sachan}$^{\alpha}$
  \quad \textbf{Bernhard Schölkopf}$^{\,\alpha, \beta}$
  \quad \textbf{Ryan Cotterell}$^{\alpha}$\\
  $^\alpha$ETH Zürich \quad $^\beta$MPI for Intelligent Systems, Tübingen \quad 
  $^\gamma$Purdue University \\
  \texttt{\{\href{mailto:andreas.opedal@inf.ethz.ch}{andreas.opedal}, \href{mailto:ryan.cotterell@inf.ethz.ch}{ryan.cotterell}\}@inf.ethz.ch} 
}

\begin{document}

\maketitle

\begin{abstract}
Many applications of large language models (LLMs) require deductive reasoning, yet models frequently produce incorrect or redundant inference steps. We frame natural language inference as a search problem where the final answer is the valid proof itself, requiring a reasoning procedure in which intermediate inferences are correct. Specifically, we investigate whether LLMs can learn to generate correct and efficient proofs with guidance from A* search---an algorithm that guarantees an optimally efficient path to a goal. We explore two training techniques: supervised fine-tuning on execution traces from A* and reinforcement learning with A*-informed process reward models. Empirically, we find that Llama-3.2 models in the 1B--3B range benefit substantially from A* post training, going from near-zero accuracy to outperforming DeepSeek-V3.2---a much larger model. Our analysis uncovers a trade-off: while simple correctness rewards maximize accuracy, A*-informed signals strike a balance between accuracy and efficiency. Furthermore, we find that on larger search spaces, models trained with imperfect heuristics exhibit superior accuracy. Our results demonstrate a promising direction towards reasoning guided by principles derived from classical search algorithms.
\end{abstract}

\section{Introduction}

Reasoning\footnote{We acknowledge that ``reasoning'' is a loaded term riddled with ambiguity (e.g., \citealp{kambhampati2024can,hoyer2025notion,kambhampati2025stop}). In this paper, we adopt a less broad but well established notion of reasoning as logical deduction under some formal system \citep[\S \rom{3}]{russelnorvig2021}.} has emerged as one of the most prominent use cases of modern large language models (LLMs).
Chain-of-thought prompting \citep{wei2022cot} first demonstrated that providing user-specified reasoning templates can dramatically improve LLM performance. Subsequently, the field shifted towards training models to produce such reasoning templates autonomously, most recently by means of reinforcement learning with verifiable rewards (RLVR; \citealp{guo2025deepseek}).
Models trained under this paradigm have achieved striking results on reasoning-based tasks such as solving competition mathematics \citep{luong-etal-2025-towards} and graduate-level science questions \citep{rein2024gpqa}.\looseness=-1

Yet, LLMs remain highly flawed reasoners in several respects: While they may often arrive at the correct final answer, they frequently generate reasoning chains that do not faithfully reflect the process by which they arrive at that answer \citep{turpin2023language,arcuschin2025chainofthought}, show sensitivity to perturbations of a problem's surface-level features \citep{jiang-etal-2024-peek,mirzadeh2024gsmsymbolic}, fail on rigorous proof writing \citep{mahdavi2025brains,petrov2025proofbluffevaluatingllms}, struggle to generalize to complex problems \citep{dziri2023faith,kordi-etal-2026-revisiting}, and exhibit human-like reasoning biases \citep{lampinen2024content,opedal2024language}. 
Beyond correctness, another pressing concern is \emph{efficiency}: models routinely generate reasoning chains that contain unnecessary inference steps \citep{chen2025think23overthinkingo1like,opedal2025efficientreasoners}. 
This matters practically---such redundant detours reduce interpretability and undermine the trust that a human evaluator would place in an agent's proof.
It also matters scientifically, since inefficient reasoning poses a scalability barrier as task complexity increases (cf. \citealp{lawsen2025commentillusionthinkingunderstanding}).\looseness=-1

\begin{figure}
    \vspace{-3pt}
    \centering
    \makebox[\textwidth][c]{%
        \hspace*{0\textwidth}%
        \includegraphics[width=1.01\textwidth]{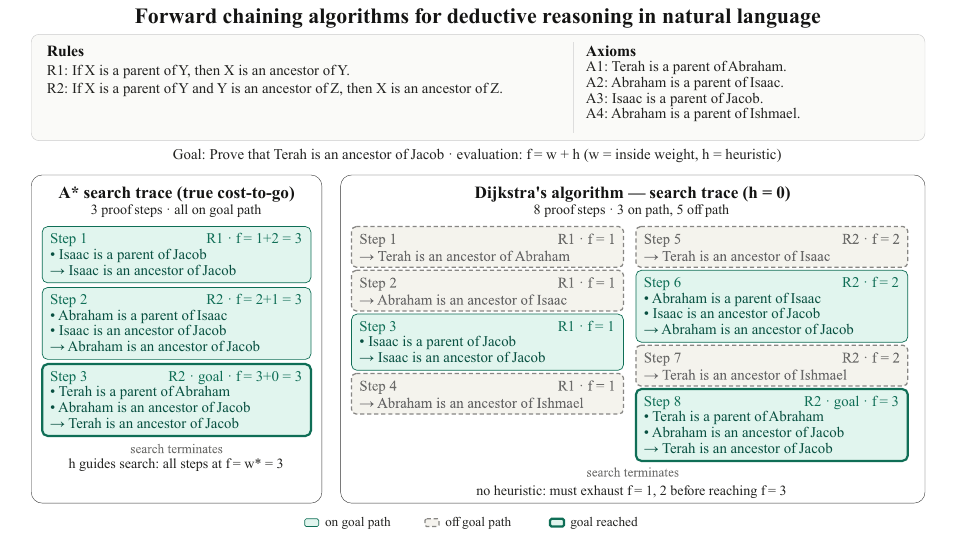}
    }
    \vspace{-14pt}
    \caption{We study whether LLMs can learn to reason correctly and efficiently in natural language under guidance from A* search. A* can yield efficient search in practice, as shown in this illustration.
    }
    \vspace{-7pt}
    \label{fig:detailed-traces}
\end{figure}

The aforementioned shortcomings are unsurprising given the dominant paradigm of training LLMs to maximize rewards based solely on final-answer correctness \citep{shao2024deepseekmathpushinglimitsmathematical}, which provides no learning signal for the correctness or efficiency of intermediate steps. A promising remedy is to train the model to not only produce correct answers, but to learn the underlying search process \citep{gandhi2024stream,gandhi2025cognitive,sel2024aot,xiang20252reasoningllmslearning}---a paradigm known as \emph{learning to search} \citep{daume2005learning,pmlr-v15-ross11a}. 
Learning to search in a principled way can yield benefits for both correctness and efficiency, but only if the model is given an appropriate learning signal.\looseness=-1 

In this paper we explore one natural source of such a signal: A* search \citep{hart1968formal}. A* is a classical path-finding algorithm which, when provided with an appropriate heuristic estimate of the cost-to-go, yields optimally efficient forward search \citep{dechter-pearl-1985-astar} with strong empirical performance on planning tasks \citep{russelnorvig2021}. 
We apply A* search to the problem of deductive reasoning, where the task is to generate a proof of a given theorem under specified logical rules and axioms, as formalized by logic programs.\footnote{Logic programming captures a broad and natural class of polynomial-time deductive reasoning tasks: \citet{Vardi1982} showed that the data complexity of evaluating Datalog queries under least fixpoint semantics is complete for \textsc{P}.} Importantly, a candidate solution is only given credit if the proof is entirely correct---since a proof that has errors is incorrect, there is \emph{no} correct final answer short of a correct proof. Employing signals from our A* implementation, we train LLMs to efficiently generate correct proofs in natural language.
We consider two approaches: (1) imitation of search traces generated by execution of A* using supervised fine-tuning (SFT) and (2) reinforcement learning (RL) with the GRPO objective \citep{shao2024deepseekmathpushinglimitsmathematical}, guided by process reward models (PRMs; \citealp{lightman2024lets}) that are derived from the A* algorithm.

We train Llama models in the 1B--3B range and find that they benefit substantially from A*-informed post-training. For example, Llama-3.2-3B-Instruct goes from $\approx 0\%$ accuracy to at best $93.5\%$ accuracy and $97.5\%$ efficiency on ProofWriter \citep{tafjord-etal-2021-proofwriter}, surpassing DeepSeek-V3.2 ($79.6\%$ accuracy, $94.3\%$ efficiency) with in-context learning. In general, SFT produces considerable gains that are further amplified by RL. We also find that search heuristic has a significant impact on performance: 
On ProofWriter data, which has a relatively small search space, the model learns best using A* coupled with the true cost-to-go. On a second dataset with a larger search space \citep{rameshkumar-etal-2025-reasoning}, however, the model achieves higher accuracy on held-out data when trained with an imperfect A* heuristic. 
We consider various reward models: a simple correctness reward, an efficiency reward based on step counts, and A*-informed rewards. 
In so doing, we uncover a trade-off between accuracy and efficiency---all reward models appear on the accuracy-efficiency Pareto front. Overall, our paper demonstrates a promising path toward generating correct and efficient proofs under the guidance of classical planning algorithms.

\section{Deductive Reasoning and A* Search}\label{sec:background}
We provide background on deductive reasoning under logic programs (\cref{sec:deductive}) and the A* search algorithm (\cref{sec:a-star}). A* was originally conceived as an algorithm over standard weighted graphs; however, we require its extension to \emph{hyper}graphs \citep{felzenszwalbmcallester2007generalized} in order to account for the fact that inference rules may have more than one premise, such as R2 in \cref{fig:detailed-traces}.

\subsection{Deductive Reasoning Formalized as Logic Programming}
\label{sec:deductive}

We consider a notion of logic programming based on Datalog \citep{ceri1989datalog}, which captures a broad class of deductive reasoning tasks computable in polynomial time \citep{Vardi1982,immerman1986relational}. Logic programming is an established framework for studying LLM reasoning (e.g., \citealp{clark2021soft,zhang2023paradox}). This subsection introduces the necessary background and notation.

An \defn{atom} is an expression of the form $\predicatep(\atom_1, \ldots, \atom_N)$, where $\predicatep$ is called a \defn{predicate} and the arguments $\atom_1, \ldots, \atom_N \in \constants\cup\variables$ are called \defn{terms}. Terms can be constants (\defn{ground} terms), denoted $\const{x}, \const{y}, \const{z}$, or variables (\defn{nonground} terms),  denoted $\varx, \vary, \varz$.
An \defn{inference rule} is an expression of the form $\premise_1, \ldots, \premise_N \vdash \conclusion$, where $\premise_1, \ldots, \premise_K, \conclusion$ are atoms; $\premise_1, \ldots, \premise_K$ are called \defn{premises} and $\conclusion$ is the \defn{conclusion}.
For example,\looseness=-1 
\begin{equation*}
    \predicate{parent}(\varx, \vary),\,\predicate{ancestor}(\vary, \varz) \vdash \predicate{ancestor}(\varx, \varz) 
\end{equation*}
is an inference rule that allows us to conclude that if $\varx$ is a parent of $\vary$ and $\vary$ is an ancestor of $\varz$, then $\varx$ is an ancestor of $\varz$.
We assume that the inference rules are \defn{range restricted}, meaning that each variable appearing in the conclusion $\conclusion$ also appears in at least one atom $\premise_k$ in the premise.
Inference rules with a null premise, i.e., where $K=0$, and a ground conclusion are called \defn{axioms}. 
For example, $\predicate{parent}(\const{abraham}, \const{isaac})$, omitting the $\vdash$ symbol, is an axiom.
A \defn{logic program} $\program$ is simply a set of inference rules, often written $\program = \rules \sqcup \axioms$ to separate axioms $\axioms$ from other rules $\rules$. For example,
\begin{align*}
     & \predicate{parent}(\const{terah},\const{abraham}) \quad \predicate{parent}(\const{abraham},\const{ishmael}) \\ 
     & \predicate{parent}(\const{abraham}, \const{isaac})  \quad \predicate{parent}(\const{isaac}, \const{jacob}) \vspace{-8pt} \\
     & \predicate{parent}(\varx, \vary) \vdash \predicate{ancestor}(\varx, \vary) \quad \predicate{parent}(\varx, \vary),\,\predicate{ancestor}(\vary, \varz) \vdash \predicate{ancestor}(\varx, \varz)  
\end{align*}
is a simple logic program that expresses biblical ancestries \citep{artofprolog1994}. A program $\program$ induces a \defn{Herbrand base} $\herbrandbase$, which is the set of all atoms that can be formed by predicates and ground terms that occur in $\program$. A subset of the Herbrand base $\interpretation \subseteq \herbrandbase$ is called an \defn{interpretation}.

We require some additional definitions to assign semantics to a logic program, i.e., to determine what is true. 
A \defn{substitution} $\subst$ is a finite set of pairs $\{(\var{X}_m, \atom_m)\}_{m=1}^M$, where
$\var{X}_m \in \variables$, $\atom_m \in \constants\cup\variables$, $\var{X}_m \neq \var{X}_{m'}$ for all $m \neq m'$, and $\var{X}_m \neq \atom_m$ for all $m$.
Substitutions $\subst$ can be applied to atoms $\premise$, e.g.,
$\predicate{parent}(\varx, \const{isaac})/\{(\varx, \const{abraham})\} = \predicate{parent}(\const{abraham}, \const{isaac})$. 
Let $\substs(\program)$ denote the set of all substitutions under $\program$.
We can write $\premise'_1, \ldots, \premise'_K \vdash \conclusion'$ if there exists a
substitution $\subst \in \substs(\program)$ such that $\premise'_1, \ldots, \premise'_K \vdash \conclusion' = \premise_1/\subst, \ldots, \premise_K/\subst \vdash \conclusion/\subst$; this is called an \defn{instantiation} of $\premise_1, \ldots, \premise_K \vdash \conclusion$.
A \defn{ground instantiation} is an instantiation that does not contain variables. We will generally use $\premise'_1, \ldots, \premise'_K, \conclusion'$ as shorthand for the atoms in ground instantiations of $\premise_1, \ldots, \premise_K \vdash \conclusion$.
Now, a program $\program$’s \defn{fixpoint operator}
$\fixpoint \colon 2^{\herbrandbase} \to 2^{\herbrandbase}$ is defined as
\begin{equation}\label{eq:fixpoint}
\fixpoint(\interpretation)
  = \interpretation \cup
(
  \{
  \conclusion/\subst \mid
  (\premise_1, \ldots, \premise_K \vdash \conclusion) \in \program,\;
  \wedge_{k=1}^K \premise_k/\subst \in \interpretation,\;
  \subst \in \substs(\program)
\}
  \cap \herbrandbase),
\end{equation}
where $\interpretation \subseteq \herbrandbase$ is an interpretation. This operator allows us to assign semantics:
The \defn{minimal Herbrand model} is $\worldmodel = \fixpointkleene(\axioms) \defeq \bigcup_{n=0}^\infty \fixpointn(\axioms)$, where $\fixpointn$ denotes the $n$-fold application of the fixpoint operator $\fixpoint$ and $\fixpointzero\defeq \axioms$. That is, $\worldmodel$ is the least fixpoint of $\fixpoint$ applied to $\axioms$ and it expresses all atoms that are true under the program;\footnote{The least fixpoint exists since for all $\interpretation_1, \interpretation_2 \subseteq \herbrandbase$, we have $\interpretation_1 \subseteq \interpretation_2 \Longrightarrow \fixpoint(\interpretation_1) \subseteq \fixpoint(\interpretation_2)$ \citep{tarski1955lattice}.}
we refer to those atoms as \defn{theorems}. In this paper, we are interested in the task of showing that a given atom $\goal$ is a theorem, i.e., to \emph{prove} $\goal$. \cref{fig:detailed-traces} gives an illustration of the problem setup in natural language.

\subsection{Theorem Proving as A* Search over a Hypergraph}\label{sec:a-star}

A* search \citep{hart1968formal} is a generalization of \citeposs{dijkstra1959note} algorithm for finding shortest paths in weighted graphs that, in addition to the cost-so-far, incorporates an estimate of the true cost-to-go.
In this section, we explain how the process of proving theorems (as defined above) can be viewed as search in a \emph{hyper}graph \citep{klein-manning-2001-parsing} and how the A* algorithm extends to this setting. Our presentation of A* is subsumed by the generalized framework of \citet{felzenszwalbmcallester2007generalized} and is analogous to A* parsers for context-free grammars \citep{klein-and-manning2002fast,klein-manning-2003-parsing}.\looseness=-1

\paragraph{Hypergraphs and Proofs.} We associate each program $\program = \rules \sqcup \axioms$ with a \defn{B-hypergraph} \citep{GALLO1993177}, which is a $(\vertices, \edges)$ tuple where $\vertices$ is a set of vertices and $\edges \subseteq 2^{\vertices} \times \vertices$ is a set of \underline{b}ackwards-hyperedges, each of the form $T \rightarrowtail \hyperhead$, where $T = \{\premn{1}, \ldots, \premn{K}\}   \subseteq \vertices$ is called a \defn{tail} and $\hyperhead \in \vertices$ is called a \defn{head}. Additionally, we label each vertex with an element of the minimal Herbrand model $\worldmodel$ using a \defn{labeler} $\labeler\colon \vertices \rightarrow \worldmodel$.
The vertex labelings must conform to constraints imposed by the axioms and rules of the program: (1) for every vertex $\vertex$ with no incoming hyperedge we require that the label is an axiom, i.e., $\labeler(\vertex) \in \axioms$, and (2) for every hyperedge $\edge = \{\premn{1}, \ldots, \premn{K}\}  \rightarrowtail \hyperhead \in \edges$, there must exist a rule $(\premise_1, \ldots, \premise_K \vdash \conclusion)\in \rules$ and a substitution $\subst \in \substs(\program)$ such that $\premise_1 / \subst = \labeler(\premn{1}), \ldots, \premise_K / \subst = \labeler(\premn{K})$ and $\conclusion / \subst = \labeler(\hyperhead)$. That is, search \emph{starts} from the axioms and \emph{proceeds} by traversing the hypergraph through rule applications based on known theorems. 
To prove a goal theorem $\goal$, the search objective is to find a \defn{hyperpath} from a source set \(\source \subseteq \labeler^{-1}(\axioms)\) (i.e., a subset of the axioms) to a goal vertex \(\goalvertex \in \vertices\) labeled with $\goal$, which is a sequence of hyperedges
\(T_1 \rightarrowtail \hyperheadj{1}, \ldots, T_J \rightarrowtail \hyperheadj{J}\)
such that \(\hyperheadj{J} = \goalvertex\) and, for every \(j \in [J]\), $T_j \subseteq \source \cup \{\hyperheadj{1}, \ldots, \hyperheadj{j-1}\}$; we refer to such a hyperpath $\proofg$ as a \defn{proof} of $\goal$.\looseness=-1

\paragraph{Shortest Proof and Weights.} Let $\realspos$ be the nonnegative reals extended with $\infty$. We associate a \defn{weight} $\weight\colon \worldmodel \rightarrow \realspos$ with each vertex label, defined recursively as follows: (1) $\weight(\premise')\defeq 0$ for all axioms $\premise' \in \axioms$ and (2) $\weight(\conclusion') \defeq\min_{\premise'_1, \ldots, \premise'_K \vdash \conclusion'} \cost(\weight(\premise'_1), \dots, \weight(\premise'_K))$ for all $\conclusion' \in \worldmodel\setminus\axioms$, where $\cost\colon\realskpos \rightarrow \realspos$ is a rule-specific \defn{cost function}.
The goal of search is to find the size $\goalweight$ of the shortest proof of a goal $\goal$, which is equal to the weight $\weight(\goal)$. There exist many notions of proof size, e.g., depth, number of vertices, or more general rule-specific aggregations; which notion is used is directly determined by the choice of cost functions.
In this work, for a rule $\premise_1, \ldots, \premise_K \vdash \conclusion$ we set 
\begin{equation}\label{eq:max_cost}
    \cost(\weight(\premise'_1), \dots, \weight(\premise'_K)) \defeq 1+ \max_{k=1}^K\weight(\premise'_k),
\end{equation}
for all corresponding ground instantiations $\premise'_1, \ldots, \premise'_K \vdash \conclusion'$. This yields a notion of size as depth, so the shortest proof will be the one with the shallowest depth (see \citealp[\S 2B]{knuth1977dijkstra}).
We give more details on cost functions and compare our choice to another plausible one in \cref{sec:cost-function}.

\paragraph{A* Search.} A* search is a forward chaining algorithm in which the ordering of expansions is determined by a particular priority queue. 
Given a program $\program = \rules \sqcup \axioms$ and a goal theorem $\goal$, it returns the size $\goalweight$ of the shortest proof of $\goal$, or $\infty$ if $\goal$ cannot be proved.
The priority of an item $\conclusion'$ depends on both the cost-so-far and the cost-to-go through the following \defn{priority function}:
\begin{equation}\label{eq:evaluation-function}
    \evaluation(\conclusion') = \weight(\conclusion') + \heuristic(\conclusion'),
\end{equation}
where 
the \defn{heuristic function} $\heuristic\colon \worldmodel \rightarrow \realspos$ estimates the true cost-to-go to the goal from $\conclusion'$. Formally, the true cost-to-go is the \defn{outside weight} $\outside(\conclusion')$ \citep{baker1979trainable,LARI199035}, defined as the value $\outside(\conclusion')$ such that $\weight(\conclusion') + \outside(\conclusion') = \goalweight$.
In order for A* search to be correct, the heuristic cannot overestimate the outside weight, i.e., $\heuristic(\conclusion') \leq \outside(\conclusion')$ for all $\conclusion'\in\worldmodel$---this property is called \defn{admissibility}. Setting $\heuristic(\conclusion')=0$ for all $\conclusion'$ corresponds to uninformed search and recovers Dijkstra's algorithm.\looseness=-1

We give pseudocode for A* search in \cref{alg:A-star}.
The algorithm maintains a priority queue $\agenda$ over theorems---called the \defn{agenda}---and a set $\chart$ of theorems---called a \defn{chart}. Two operations will be particularly important for the remainder of this paper: \emph{pushes} and \emph{pops}.
Each iteration \emph{pops} the item $\premise'$ from $\agenda$ with the lowest priority $\evaluation(\premise')$ according to \cref{eq:evaluation-function}, adds it to the chart, and checks whether new rules can be fired combining $\premise'$ with the items existing in the chart.
Any new conclusions are \emph{pushed} to $\agenda$ along with their appropriate priorities. The algorithm terminates on popping the goal item.\footnote{We have omitted backpointers from the pseudocode, which are used to return the shortest proof itself in addition to its size. Our code implementation also returns the trace of logical inferences that were made during search, as will be discussed in \cref{sec:verbalizing}.}\looseness=-1

\begin{algorithm}[t]
\footnotesize
\caption{Generalized A* Search for Deductive Reasoning over Hypergraphs}
\label{alg:A-star}
\begin{algorithmic}[1]
\Function{\textsc{A*}}{$\program = \rules \sqcup \axioms, \goal$}
\LineComment{Program $\program = \rules \sqcup \axioms$; goal theorem $\goal$}
\State{$\chart \leftarrow \varnothing; \agenda \leftarrow \varnothing$}
\Comment{initialize the chart $\chart$ (a set) and the agenda $\agenda$ (a priority queue)}
\For{$\premise' \in \worldmodel$} $\weight(\premise') \leftarrow \infty$
\Comment{default item weights are $\infty$ (not yet derived)}\EndFor
\For{$\premise' \in \axioms$} 
\Comment{push axioms to agenda}
\State{$\weight(\premise') \leftarrow 0$}
\Comment{axioms have weight $\weight$ of zero}
\State{$\agenda\textsc{.push}\!\bigl(\premise',\; \heuristic(\premise')\bigr)$}
\Comment{axiom priorities are only based on heuristic values $\heuristic$}
\EndFor
\While{$\agenda \neq \varnothing$}
    \State{$\premise' \leftarrow \agenda\textsc{.pop}()$}
    \Comment{pop the item with lowest priority (smallest value of $\cost(\cdot, \dots, \cdot)+\heuristic(\cdot)$)}
    \If{$\premise' \in \chart$} \textbf{continue}
    \Comment{the item has already been processed}
    \EndIf
    \State{$\chart \gets \chart \cup \{\premise'\}$}
    \Comment{add item to chart}
    \If{$\premise' = \goal$} \textbf{return} $\weight(\goal)$
        \LineComment{the algorithm terminates when the goal is popped; correctness ensures that $\weight(\goal)=\goalweight$}
    \EndIf
    \For{$(\premise_1, \ldots, \premise_K \vdash \conclusion) \in \rules$}
        \Comment{check whether $\premise'$ has unlocked new rule applications}
        \For{$\subst \in \substs(\program)\colon \premise_1/\subst, \ldots, \premise_K/\subst, \conclusion/\subst \in \herbrandbase\ \textbf{and}\ \forall k\in\{1,\dots,K\}\colon \premise_k/\subst \in \chart$}
        \Comment{check for premises in $\chart$}
            \State{$\conclusion' \gets \conclusion/\subst$}
            \Comment{new conclusion is derived}
            \State{$\weight' \gets \cost(\weight(\premise_1/\subst), \dots, \weight(\premise_K/\subst))$}
            \Comment{compute the weight of new conclusion}
            \If{$ \conclusion' \notin \chart\ \textbf{and}\ \weight' < \weight(\conclusion')$}
                \Comment{new weight is cheaper than known weight}
                \State{$\weight(\conclusion') \gets \weight'$}
                \Comment{update the weight}
                \State{$\agenda\textsc{.push}\!\bigl(\conclusion',\; \weight' + h(\conclusion')\bigr)$}
                \Comment{push to the agenda}
            \EndIf
        \EndFor
    \EndFor
\EndWhile
\State\Return{$\weight(\goal)$}
\Comment{goal not provable; the weight is $\infty$}
\EndFunction
\end{algorithmic}
\end{algorithm}

\paragraph{Guarantees on Correctness and Efficiency.}
A heuristic function $\heuristic$ is \defn{consistent} (or \defn{monotone}) if for all $(\premise_1, \ldots, \premise_K \vdash \conclusion) \in \rules$ and $\subst \in \substs(\program)$ with $\premise'_k = \premise_k/\subst$ and $\conclusion' = \conclusion/\subst$ such that $\premise'_1, \dots, \premise'_K, \conclusion' \in \worldmodel$,
\begin{equation}\label{eq:consistency}
    \weight(\premise'_k) + \heuristic(\premise'_k) \leq \cost(\weight(\premise'_1), \ldots, \weight(\premise'_K)) + \heuristic(\conclusion'), \quad \forall k = 1, \ldots, K,
\end{equation}
for all weight assignments $\weight(\premise'_1), \ldots, \weight(\premise'_K)$ possible under the algorithm. Intuitively, this means that the priority function $\evaluation$ is monotonically nondecreasing along any hyperpath. Consistency implies admissibility if $\heuristic(\goal)=0$ \citep{felzenszwalbmcallester2007generalized}.
If $\heuristic$ is consistent, it follows that for all items $\conclusion'$, the derived weight equals the true weight when $\conclusion'$ is added to the chart at line 11 in \cref{alg:A-star} \citep[Thm 2]{felzenszwalbmcallester2007generalized}. This implies correctness: the weight $\weight(\goal)$ returned by \cref{alg:A-star} is the weight $\goalweight$ of the shortest proof of $\goal$.
In addition, given a consistent heuristic function, A* is \defn{optimally efficient} in the sense that it pops the fewest number of items (line 9 in \cref{alg:A-star}) required to prove the goal by any algorithm with access to the same heuristic information \citep{dechter-pearl-1985-astar}. 
The choice of heuristic function plays a significant role for efficiency in practice, which our experiments explore in the context of LLM reasoning (\cref{sec:experiments}).

\section{From Symbolic Search To Language Model Reasoning}\label{sec:llm-reasoning}
The previous section presented background on A* search under a symbolic framework for deductive reasoning. In this section we bridge the gap between reasoning under logic programs and reasoning with natural language tokens as generated by LLMs (\cref{sec:verbalizing}). We then discuss how we derive signals from A* search for reward modeling (\cref{sec:reward-modeling}).

\subsection{Representing Deductive Search in Natural Language}\label{sec:verbalizing}

LLMs reason with strings over natural language tokens. We therefore require a way of mapping the execution of the algorithm into a natural language string. We discuss our approach and contrast it to \citet{lehnert2024beyond}, who trained transformers from scratch on A* traces for classical planning tasks.

\paragraph{Tracing the Execution.} How can the execution of \cref{alg:A-star} be represented as a sequence? \citet{lehnert2024beyond} propose one answer: write out the items that were popped (line 9 in \cref{alg:A-star}) and pushed (lines 7 and 20 in \cref{alg:A-star}), along with corresponding values of $\weight$ and $\heuristic$ (\cref{eq:evaluation-function}), following the order in which they occur during execution. However, while they append the proof returned by the algorithm when creating training data, there is generally no way of determining the proof based solely on their notion of execution trace, without additional bookkeeping. In particular, when pushing an item $\conclusion'$, it is necessary to know the tail vertices $\premise'_1, \dots, \premise'_K$ from which $\conclusion'$ was derived---so-called \defn{backpointers}. 

Our goal is to verify that an LLM's proof is correct, which requires stating \emph{why} each theorem in the proof is true.
We therefore propose an alternative to \citet{lehnert2024beyond} where we represent each push as a \defn{proof step}, which includes the necessary backpointers and the inference rule that was applied. Formally, for a given (program, goal)-pair $(\program,\goal)$, with substitutions $\substs$ and minimal Herbrand model $\worldmodel$ (see \cref{sec:deductive} for definitions), consider the following alphabet of symbols:
\begin{equation}
    \alphabet \defeq \{ \textsc{push}((\premise_1/\subst, \dots, \premise_K/\subst), \arule, \conclusion/\subst) \mid \premise_1/\subst, \dots, \premise_K/\subst, \conclusion/\subst \in \worldmodel, \arule \!=\!(\premise_1, \dots, \premise_K \vdash \conclusion)\!\in \rules, \subst \in \substs \}.
\end{equation}
That is, $\alphabet$ contains symbols corresponding to all valid proof steps. (Note that $\alphabet$ is countably infinite for infinite minimal Herbrand models.) Let $\kleene{\alphabet}$ be the Kleene closure of $\alphabet$, i.e., the set of all strings formed by symbols in $\alphabet$. We consider a \defn{search trace} $\trace$ to be an element $\trace \in \kleene{\alphabet}$ such that the constraints of \cref{alg:A-star} are satisfied, i.e., the symbols in $\trace$ correspond to pushes executed by the algorithm in the right order. These traces are readily obtained by additional bookkeeping in \cref{alg:A-star}.

\paragraph{Pushes and Pops.} A search trace as we have defined it does not explicitly contain pops, but it does carry information about them: For a step $\textsc{push}((\premise'_1, \dots, \premise'_K), \arule, \conclusion')$, we know that all items $\premise'_1, \dots, \premise'_K$ must have been popped because \cref{alg:A-star} requires that they all be in the chart. We are not concerned with the actual pop ordering of $\premise'_1, \dots, \premise'_K$ since this ordering does not matter for whether a rule can be applied.
That is, as long as all premises are popped before the conclusion, we are agnostic to which ordering the LLM may have popped the premises. 
We can derive the minimum number of items that must have been popped to yield a given search trace by counting the unique premises that appear in the search trace, i.e., the number of elements $|\popset(\trace)|$ in the set
\begin{equation}
    \popset(\trace) \defeq \!\!\!\!\!\bigcup_{\textsc{push}((\premise'_1, \dots, \premise'_K), \arule, \conclusion') \in \trace} \!\!\!\!\!\{\premise' \mid \premise' \in \{\premise'_1, \dots, \premise'_K\}\}.
\end{equation}
This is the \emph{minimum} number of required pops since there may have been additional pops that did not result in any push. 
Empirically, we evaluate efficiency both in terms of proof steps (pushes) and (minimal) pops; the former relates more closely to the token length of the search trace, while the latter represents what the A* search algorithm optimizes for; see \cref{sec:a-star}.
Our notion of search trace can be formalized as a quotient set induced by a string homomorphism over traces that additionally contain pops. In layman's terms: What we consider is an equivalence class on pop orderings that yield the same push ordering. See \cref{sec:equivalence-class} for details and a more formal comparison to \citet{lehnert2024beyond}.\footnote{\citet{lehnert2024beyond} additionally included $\weight$ and $\heuristic$ in the search trace. We do not require the LLM to generate these numbers explicitly, only to maintain some internal representation of them. See \cref{sec:related-work} for a more general discussion of related work.\looseness=-1}

\paragraph{Verbalized Search Traces.} Let $\lmalphabet$ be an alphabet of natural language tokens. A \defn{language model} $\lm$ is a probability distribution over the Kleene closure of $\lmalphabet$---denoted $\kleene{\lmalphabet}$. 
To bridge the search trace with natural language we associate each atom and rule with a set of strings. For a program $\program$, we define a \defn{verbalizer}
$\generator{\program} \colon \herbrandbase^* \cup \rules \rightarrow 2^{\kleene{\lmalphabet}}$, where $\herbrandbase^*$ is the Kleene closure of $\herbrandbase$ and each $\generator{\program}(\premise')$ must be a disjoint set. A search trace is verbalized by applying $\generator{\program}$ to all atoms and rules in each of the proof steps in the order as they occur, separating the proof steps by a distinguished delimiter sequence---in our case ``\textbackslash n\textbackslash n''. Axiom verbalizations are included in the prompt and are not repeated in the verbalized search trace. See \cref{sec:prompt-verbalizations} for an example on how this is done in our experiments.
\cref{fig:detailed-traces} gave an informal example of verbalized search traces from A* with the true cost-to-go and Dijkstra's algorithm.\looseness=-1

\subsection{Deductive Search in Natural Language as a Reinforcement Learning Problem}\label{sec:reward-modeling}

Our aim is to train LLMs to generate correct and efficient verbalized search traces. This is an instantiation of the learning to search paradigm \citep{daume2005learning, Daume2009search, pmlr-v15-ross11a}, applied to natural language reasoning. A straightforward learning-to-search approach is to train a predictor based on expert demonstrations, which corresponds to supervised fine-tuning (SFT) for LLMs. Such form of imitation learning, however, may lead to a compounding of errors if the observations at inference differ from those in the expert demonstration \citep{pmlr-v9-ross10a}.\looseness=-1

Therefore, we also consider a reinforcement learning (RL) approach. The verbalized search trace provides dense signals with regard to whether the proof steps taken were correct and whether the search trace corresponds to the execution of an efficient algorithm. These dense signals can be quantified through the priority function of A* search from \cref{eq:evaluation-function}; see \citet{gehring2022reinforcement} for related ideas. We use these priority scores to assign credit through process reward models (PRMs; \citealp{lightman2024lets,setlur2025rewarding}). 
Note that we have access to ground-truth priority scores at training and inference time by parsing the verbalized search trace and calling \cref{alg:A-star}. Thus, we do not need to train a neural PRM from data; however, future work may explore using A* search traces to that end.\looseness=-1

\paragraph{A*-informed Process Rewards.} Let $(\program,\goal)$ be a (program, goal)-pair and $\trace$ be a candidate search trace, e.g., corresponding to an LLM solution. We obtain a raw efficiency score by summing the priority values over all items that we know have been popped: $\score_{\trace} \defeq \sum_{\premise' \in \popset(\trace)} \evaluation(\premise')$,
where $\popset$ is the minimum set of popped items entailed by $\trace$ (\cref{sec:verbalizing}) and $\evaluation$ is from \cref{eq:evaluation-function}. Such raw scores are inappropriate for reward modeling, however, since their range can vary drastically across programs and goals, which could lead to unstable gradient optimization (see, e.g., \citealp{hasselt2016learning, schaul2021returnbasedscalingnormalisationtrick}). Moreover, standard RL objectives are built for maximizing returns, not minimizing costs. In our experiments, we therefore normalize the raw scores with an exponentially decaying reward model:\looseness=-1
\begin{equation}\label{eq:reward-exponential-decay}
    \reward(\trace) \defeq 
    \begin{cases}
        \exp\left(\scale\cdot \left(\location-\sum_{\premise' \in \popset(\trace)} \evaluation(\premise')\right)\right), & \text{if}\ \trace\ \text{correctly proves}\ \goal\ \text{from}\ \program\\
        0, & \text{otherwise},
    \end{cases}
\end{equation}
with scale and location parameters $\scale \in \reals_{>0}$ and $\location \in \realspos$. The intuition behind \cref{eq:reward-exponential-decay} is to give a non-zero reward if the search trace is correct; if it is, the reward is scaled based on efficiency scores as assigned by A*'s priority function.
We set $\location$ and $\scale$ adaptively based on $(\program,\goal)$. Specifically, we set $\location=\score_{\trace^*}$, where $\trace^*$ is the true search trace returned by A* on input $(\program, \goal)$. This ensures a reward of $1$ if the candidate trace is perfectly aligned with the underlying A* algorithm. Note, however, that the reward will be greater than $1$ if the output is more efficient than the underlying A* algorithm, whose efficiency depends on the heuristic function (discussed below).
We set $\scale=\log(0.5)/\score_{\trace^*}$, which results in a scaling behavior in which doubling the score compared to $\score_{\trace^*}$ yields half the reward.\looseness=-1

\paragraph{Heuristic Functions.} The above reward model is compatible with arbitrary A* heuristics, the choice of which is important for the practical efficiency of A* \citep{russelnorvig2021}. A heuristic that is closer to the true cost-to-go will generally lead to more efficient search in terms of number of pops. A common way to derive heuristics is to compute the cost-to-go for relaxed variants of the problem, in which fewer restrictions are enforced on when certain rules are applicable. 
In other words, a new hypergraph $(\vertices', \edges')$ is constructed for which the original hypergraph $(\vertices, \edges)$ is a subhypergraph, i.e., $\vertices \subseteq \vertices'$ and $\edges \subseteq \edges'$, where the cost functions associated with hyperedges in $(\vertices, \edges)$ are preserved in $(\vertices', \edges')$. The cost-to-go from the relaxed problem will be a consistent heuristic \citep{felzenszwalbmcallester2007generalized}. 
Following this idea, we adopt a heuristic based on the coarsened, unary dependencies encoded in the program's rules. For all ground instantiations $\premise'_1, \ldots, \premise'_K \vdash \conclusion'$, we add $K$ edges $\{\premn{1}\} \rightarrowtail \hyperhead, \dots, \{\premn{K}\} \rightarrowtail \hyperhead$, where $\premn{1}, \dots, \premn{K}, \hyperhead$ are the corresponding vertices in the original hypergraph. Intuitively, this heuristic encodes the dependency of the rules without considering which theorems are entailed by the axioms; we call it the \defn{dependency heuristic}. Our experiments additionally use the true cost-to-go, i.e., the outside weights, as the heuristic function.

\section{Can A* Post-Training Yield Improvements in Reasoning?}\label{sec:experiments}
This section addresses the paper's main empirical question.
\cref{sec:experimental-setup} explains the experimental setup and \cref{sec:results-sft}-\ref{sec:results-rl} discuss the results.

\subsection{Experimental Setup}\label{sec:experimental-setup}

\paragraph{Datasets.} We take logic programs and goal theorems from ProofWriter \citep{clark2021soft,tafjord-etal-2021-proofwriter} and DeepRD \citep{rameshkumar-etal-2025-reasoning}. For ProofWriter, we only consider problems where the depth of the shortest proof is at least $3$ and where the goal is provable under the program. Applying this filtering yields a dataset with $9207$ training examples, $1352$ validation examples, and $2615$ test examples. DeepRD is a programmable dataset which supports arbitrarily complex programs along two dimensions, a lookahead value $L$ determining the depth of the shortest proof and a branching factor $B$ determining the number of branches from the single axiom vertex. We generate a dataset with $2000$ training examples, $200$ validation examples, and $500$ test examples using values $L\in \{5,\dots,10\}$ and $B\in\{4,\dots,8\}$. 
\cref{fig:proof_steps_comparison_proofwriter,fig:proof_steps_comparison_deeprd} show the number of proof steps (i.e., pushes) performed by the three different search orders on ProofWriter and DeepRD, respectively. We make two remarks based on these plots: (1) the DeepRD problems are considerably longer than the ProofWriter problems and (2) the dependency heuristic performs similarly to the true cost-to-go on ProofWriter, whereas on DeepRD the dependency heuristic gives little advantage over Dijkstra's.

\paragraph{Evaluation.} 
Importantly, we evaluate the entire verbalized search trace generated by the LLM. In contrast, most prior work treat natural language theorem proving as a simplified task such as binary classification (e.g., \citealp{clark2021soft}).
We parse the LLM's output into search traces using a specified format for segmentation; \cref{sec:prompt-verbalizations} gives an example. We allow arbitrary token sequences between the blocks of verbalized proof steps and consider variations of verbalizations beyond what is in the training data.
Solutions are scored on correctness and efficiency. Correctness (\defn{accuracy}) is a binary score; a search trace is considered correct if it yields a valid proof of the goal under the axioms and rules of the program. That is, any incorrect proof step required to prove the goal will result in a score of $0$. Efficiency is scored by comparing \emph{correct} proofs against the ground-truth shortest proof of the problem, following \citet{opedal2025efficientreasoners}. We consider two variants: \defn{efficiency (pushes)}, which measures the number of proof steps (or pushes), and \defn{efficiency (pops)}, which measures the minimum number of popped items required to yield the search trace. For both metrics we take the proportion against the corresponding number in the shortest proof, resulting in a score between $0$ and $1$.

\begin{figure}[t]
    \centering
    \vspace{-10pt}
    \includegraphics[width=0.99\linewidth]{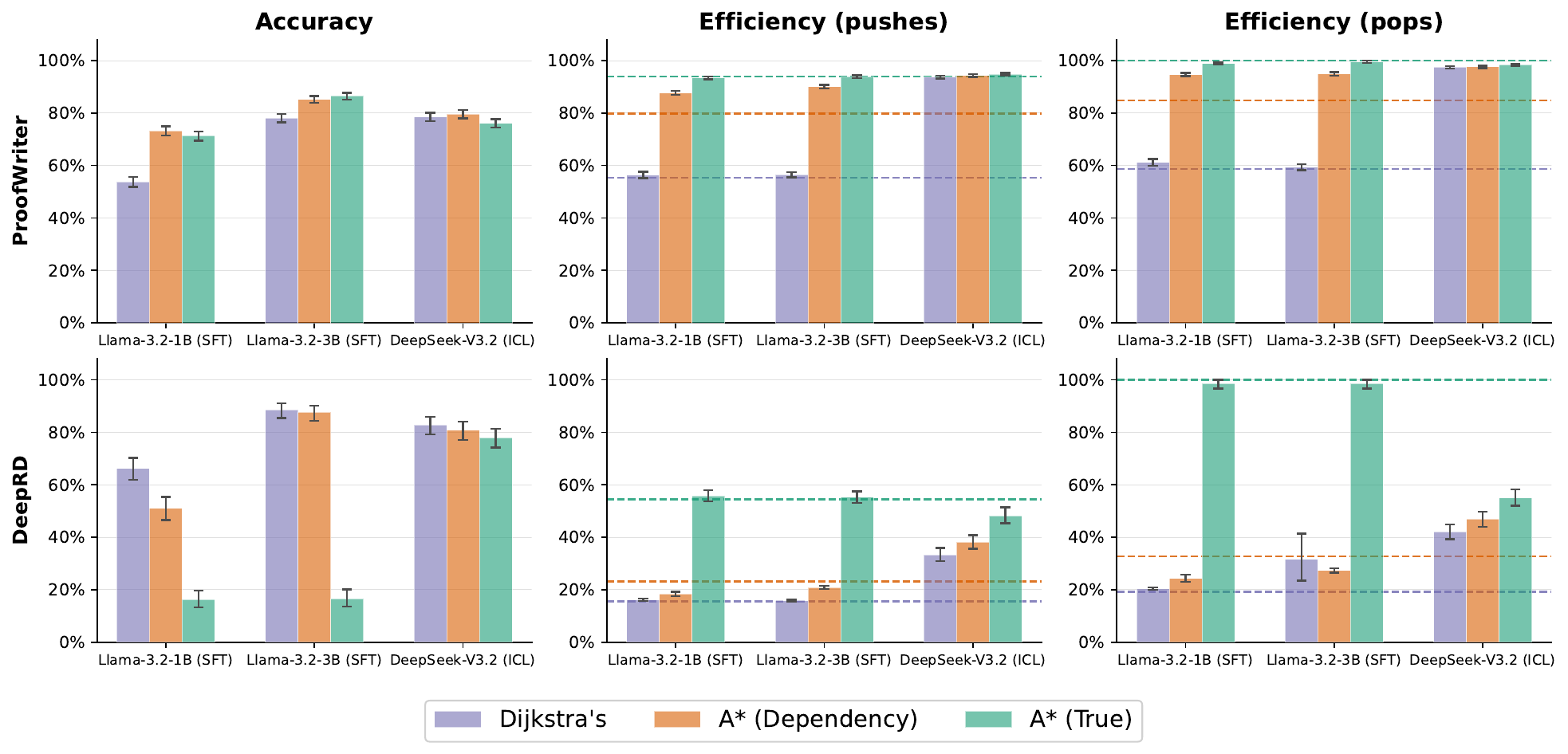}
    \vspace{-8pt}
    \caption{Performance of LLMs finetuned with SFT across different search orders, as compared to an ICL baseline with DeepSeek-V3.2, evaluated on the test sets of ProofWriter and DeepRD. 
    We observe that the performance of the finetuned models is comparable to, or outperforms, DeepSeek-V3.2.
    The dashed lines indicate the efficiency scores for the corresponding search algorithm that matches the color. Intervals are 95\% Wilson score CIs for accuracy and bootstrap CIs for efficiency. \looseness=-1}
    \vspace{-6pt}
    \label{fig:sft-results}
\end{figure}

\paragraph{Search Orders and Training.}
We finetune Llama-3.2-1B-Instruct and Llama-3.2-3B-Instruct \citep{dubey2024llama3herdmodels} using supervised fine-tuning (SFT) and RL with GRPO \citep{shao2024deepseekmathpushinglimitsmathematical} and the PRM discussed in \cref{sec:reward-modeling}. As baselines we consider in-context learning (ICL) of the base models and of DeepSeek-V3.2 \citep{deepseek3.2-2025}. For ICL and SFT we verbalize search traces according to three different search heuristics:
(i) Dijkstra's (uninformed search), 
(ii) A* with the dependency heuristic, and (iii) A* with the true cost-to-go. We feed the examples from the datasets to our implementation of \cref{alg:A-star} and translate the resulting search traces into verbalizations using existing templates from the datasets. In practice, several items often have the same priority value. Such ties are broken following recommendations by \citet{asai-fukunaga-2017-tie}, firstly by value of heuristic estimate (lower is better) and secondly by last-in-first-out.
For RL we consider four different reward models: (i) a \defn{correctness} reward that assigns $1$ if the verbalized search trace correctly proves the goal and $0$ otherwise, (ii) a \defn{step count} reward which instantiates \cref{eq:reward-exponential-decay} as an exponential decay over the number of correct verbalized proof steps, where $\alpha$ is set as the number of steps in the shortest proof, and A*-informed reward models which instantiate \cref{eq:reward-exponential-decay} with (iii) the dependency heuristic and (iv) the true cost-to-go. Additional details such as hyperparameter values are given in \cref{sec:training-details}.

\subsection{Imitation of Search with Supervised Fine-tuning}\label{sec:results-sft}

We first note that the Llama-3.2-1B and Llama-3.2-3B models are unable to perform the tasks without training---the ICL baseline accuracies are $\approx 0\%$, so we omit them from the figures. Anecdotally, we observe that many of their responses under ICL contain hallucinated rules and inferences. The results for SFT are presented in \cref{fig:sft-results}. 
When trained on the ProofWriter data, the 3B model outperforms the DeepSeek baseline with A* fine-tuning, whereas the 1B model shows comparable performance. Training on A* traces outperforms training on traces from Dijkstra's algorithm in terms of both efficiency and correctness. The choice of A* heuristic makes little difference for accuracy on this dataset, but in terms of efficiency, the true cost-to-go yields higher performance on the test set.

For DeepRD, which has a considerably larger search space than ProofWriter (see \cref{fig:proof_steps_comparison_proofwriter,fig:proof_steps_comparison_deeprd}), SFT exposes a trade-off between accuracy and efficiency. The true cost-to-go does poorly on accuracy but the problems that are solved are done so efficiently. Dijkstra's, on the other hand, generalizes best to the test set in terms of accuracy---with the 3B model beating DeepSeek-V3.2---but is restricted by inefficient search. We emphasize, however, that efficiency can only be evaluated on examples where the proof is correct. These results suggest that learning the true heuristic is hard if the search space is large. This is consistent with previous findings that learning to search outperforms training on optimal paths \citep{gandhi2024stream, lehnert2024beyond}. We expect that the accuracy based on true cost-to-go traces could be improved with more data, however.

We additionally compare efficiency against the baselines of the underlying search algorithms, as shown by dashed lines in \cref{fig:sft-results}. These scores are computed across the full test set. (Note that these algorithms are guaranteed to be correct on all examples.)
We conclude that SFT is quite effective for imitating the exact search order: for the problems that the models solve, the efficiency is quite close to that of the underlying algorithm. 
SFT on traces with the dependency heuristic, in particular, does even better than the underlying algorithm for ProofWriter. 
The performance is overall bottlenecked by accuracy rather than efficiency, which might be expected when imitating search traces that are already efficient.\looseness=-1

\subsection{Reinforcement Learning for Efficient Search}\label{sec:results-rl}

\begin{figure}[t]
    \centering
    \vspace{-10pt}
    \includegraphics[width=0.99\linewidth]{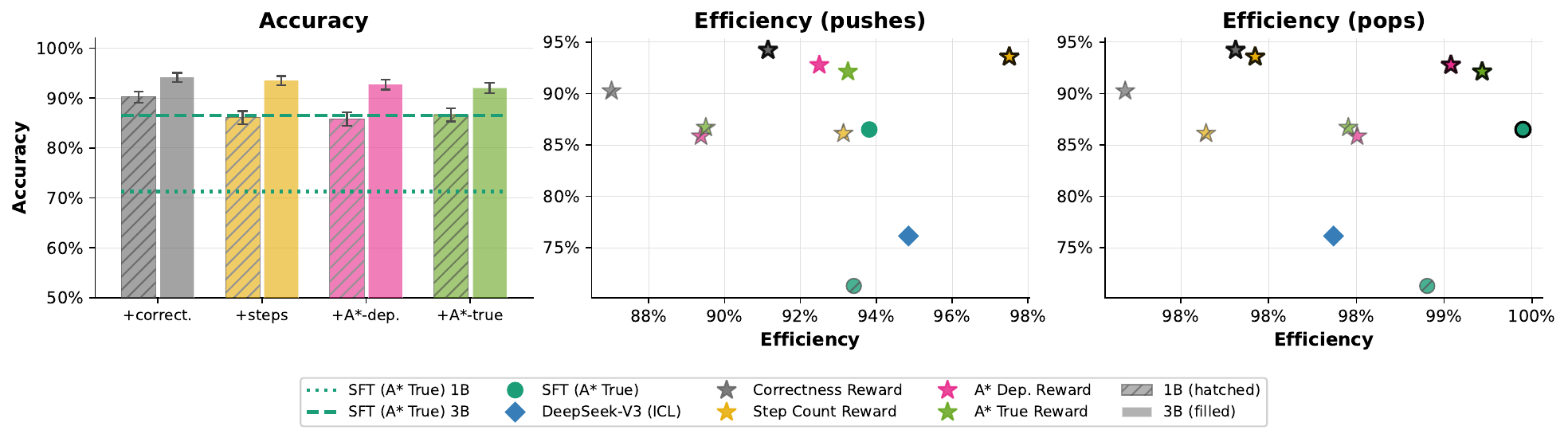}
    \vspace{-6pt}
    \caption{Results on \textbf{ProofWriter} for GRPO-finetuned models, starting from SFT checkpoints that were trained on A* traces under the \emph{true cost-to-go}. Points on the Pareto front are highlighted. We observe that all reward models are placed on at least one of the two accuracy-efficiency Pareto fronts.}
    \vspace{-6pt}
    \label{fig:3b-grpo-results}
\end{figure}

\cref{fig:3b-grpo-results} shows performance on ProofWriter data resulting from GRPO fine-tuning from the true cost-to-go SFT checkpoint with the various reward models. 
All reward models lead to improved accuracy over the corresponding SFT baseline, with the 1B models showing accuracy comparable to or even better than the SFT-tuned 3B model.
The correctness reward yields the best correctness performance overall. As expected, however, the other three rewards yield better efficiency than the simple correctness reward.  When measuring efficiency in terms of pops---which A* search optimizes for---the two A* rewards are preferable; they achieve the highest efficiency and are both on the Pareto front when comparing to other models of the same size. When measuring push efficiency, however, the step count reward is preferable. In general, no reward model dominates any of the others on both accuracy and efficiency.\looseness=-1

\begin{figure}[t]
    \centering
    \vspace{-10pt}
    \includegraphics[width=0.99\linewidth]{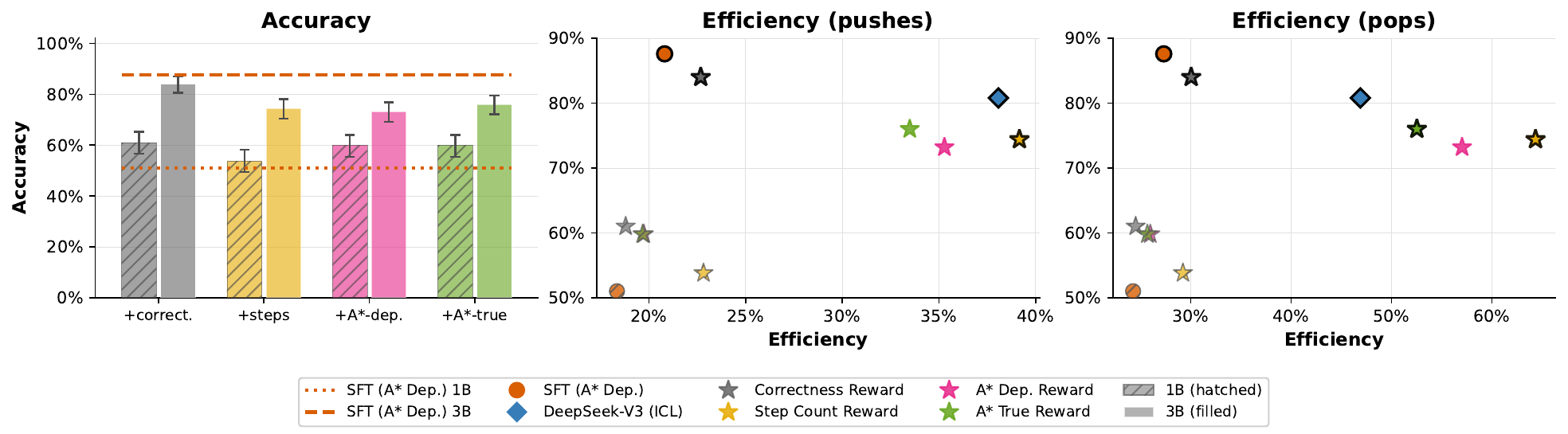}
    \vspace{-6pt}
    \caption{Results on \textbf{DeepRD} for GRPO-finetuned models, starting from SFT checkpoints that were trained on A* traces under the \emph{dependency heuristic}. Points on the Pareto front are highlighted.
    The accuracy-efficiency trade-off is shown even more starkly than in \cref{fig:grpo-proofwriter-dependency}.}
    \vspace{-6pt}
    \label{fig:grpo-deeprd-dependency}
\end{figure}

\cref{fig:grpo-deeprd-dependency} shows analogous results on the DeepRD dataset, with the difference that the SFT checkpoint is based on A* with the dependency heuristic rather than the true cost-to-go. In general, the two A* rewards strike a balance between correctness and efficiency: the step count reward yields efficient proofs but mostly lower accuracy than A* rewards, while the correctness reward yields more accurate but less efficient proofs. Interestingly, while all rewards dominate the SFT checkpoint for the 1B model, the 3B SFT model appears on the Pareto front as the most accurate option. We also note that, while the models are less competitive against DeepSeek-V3.2 on this data, DeepSeek-V3.2 does not dominate.\looseness=-1

We additionally train Llama-3.2-1B-Instruct from the dependency heuristic SFT checkpoint on ProofWriter and on the true cost-to-go SFT checkpoint on DeepRD, shown in \cref{fig:grpo-proofwriter-dependency,fig:grpo-deeprd-true} in \cref{sec:appendix-results}.

\section{Related Work}\label{sec:related-work}
\paragraph{Logical Reasoning.} Many studies have focused on measuring and improving transformer-based LLMs on deductive reasoning tasks in natural language \citep{liu2021logicqa,creswell2023selectioninference,kazemi-etal-2023-lambada,han-etal-2024-folio,wan-etal-2024-logicasker,li2025questbenchllmsaskright,madaan-etal-2025-lost}, with some existing datasets being based on logic programs \citep{clark2021soft,tafjord-etal-2021-proofwriter,zhang2023paradox}. However, the majority of these papers reduce reasoning to a simplified setting, such as binary classification of whether a statement is entailed by a program, multiple-choice question answering, or predicting the next step given a partial verbalized proof. Our introduction argued that LLMs should be evaluated based on the entire proofs; see, e.g., \citet{pmlr-v202-morishita23a} and \citet{saparov2023language} for previous work in this direction. Several papers have proposed methods to improve reasoning by incorporating symbolic solvers that are external to the LLM \citep{olausson-etal-2023-linc,pan-etal-2023-logic,quan-etal-2024-verification}. The approach by \citet{xu2026logicreward} is related to our work; they use a reward model based on a theorem prover for scoring logical validity. However, they evaluate performance in terms of final answer accuracy.

\paragraph{The ``Overthinking'' Problem.} LLMs post-trained to solve reasoning tasks with RLVR (so-called reasoning models; \citealp{guo2025deepseek}) have been found to be subject to ``overthinking''---that is, generating solution traces that are far longer than what should intuitively be required to solve the problem \citep{chen2025think23overthinkingo1like,pu2025thoughtterminatorbenchmarkingcalibratingmitigating}. Several papers have aimed to address this by incorporating rewards based on the length of the generated token sequence \citep[\emph{inter alia}]{arora2025traininglanguagemodelsreason,luo2025opruner,ayoub2026learning,liu2026learn}. This issue is related, but separate from the notion of efficiency considered in this paper: An LLM trained to find the correct final answer with few tokens may still perform incorrect proof steps or take shortcuts. Conversely, an LLM trained to find an efficient proof, e.g., using our method, may still require a long sequence of reasoning tokens to do so.\looseness=-1 

\paragraph{Search and the A* Algorithm.}
Several recent papers have advocated a learning-to-search approach to LLM reasoning \citep{gandhi2024stream,sel2024aot,shah2024causal,saparov2025transformers,xiang20252reasoningllmslearning}. We argue that it is desirable to constrain search to inferences that are \emph{correct}.
Our method is most related to \citet{lehnert2024beyond}, who found that training transformers from scratch on A* search traces outperforms solution-only training on synthetic planning tasks. We explore whether A*-based training leads to gains in a different setting: fine-tuning larger models on deductive reasoning in natural language, using both imitation learning via SFT as well as RL via A*-informed rewards.
We are unaware of previous work applying A* search to natural language reasoning, which requires an extension of A* to hypergraphs as presented in \cref{sec:a-star}. Such A* extensions have previously been applied to context-free parsing \citep{klein-manning-2003-parsing,pauls-klein-2009-k}. Other lines of work have employed A* for natural language decoding \citep{meister-etal-2020-best,lu-etal-2022-neurologic,wan-etal-2023-faithfulness} and used LLMs in combination with A* to improve search for classical planning tasks \citep{silver2022pddl,gupta-li-2024-training,meng-etal-2024-llm}.

\section{Conclusion}
Our paper framed deductive reasoning in natural language as a task where the entire proof must be correct. Building on the learning-to-search paradigm, we explored whether signals from A* search can be used to train models to prove theorems correctly and efficiently. With both supervised fine-tuning and reinforcement learning, we found that models in the 1B--3B range can be trained to be on par with, or outperform, DeepSeek-V3.2. We considered several variations of search heuristics, reward models, and datasets. Overall, we discovered a good recipe in starting with supervised fine-tuning on A* traces under a useful but imperfect heuristic, and continuing with reinforcement learning using a reward model informed by correctness and/or efficiency---the choice of which should depend on what is prioritized. Our framework, analysis, and results advance our understanding about training LLMs to perform search, demonstrating potential for methods guided by classical planning algorithms.\looseness=-1

\bibliography{bibliography}
\bibliographystyle{bibliography}

\appendix

\section{More Details on A* Search and Verbalizations}
We provide more details on A* search as applied here to hypergraphs and verbalized natural language reasoning. 
\cref{sec:equivalence-class} provides a formal discussion of the verbalized search traces and compares our method to \citeposs{lehnert2024beyond}. \cref{sec:cost-function} discusses the choice of cost function.

\subsection{Formal Comparison to \citet{lehnert2024beyond} and Equivalence over Pop Orderings}\label{sec:equivalence-class}

\cref{sec:verbalizing} mentioned how there is an equivalence class over orderings of pops resulting from execution of \cref{alg:A-star} that lead to the same ordering of pushes. This section formalizes that claim by spelling out how search traces are quotient sets induced by string homomorphisms. We compare the homomorphism considered here to the one corresponding to \citeposs{lehnert2024beyond} approach.

We fix a program $\program$ and a goal theorem $\goal$, with substitutions $\substs$ and minimal Herbrand model $\worldmodel$; we refer back to \cref{sec:deductive} for definitions. Consider the following alphabets: 
\begin{align}
    \alphabet_{\textsc{pop}} & \defeq  \{\textsc{pop}(\conclusion/\subst) \mid \conclusion/\subst \in \worldmodel\}, \\
    \alphabet_{\textsc{push}} & \defeq  \{\textsc{push}(\conclusion/\subst) \mid \conclusion/\subst \in \worldmodel\}, \\
    \alphabet_{\textsc{step}}\!\! & \defeq \{ \textsc{push}((\premise_1/\subst, \dots, \premise_K/\subst), \arule, \conclusion/\subst)\! \mid \!\premise_1/\subst, \dots, \premise_K/\subst, \conclusion/\subst \in \worldmodel, \arule \!=\!(\premise_1, \dots, \premise_K \vdash \conclusion)\!\in \rules, \subst \in \substs \}, \label{eq:push-vocab}
\end{align}
That is, $\alphabet_{\textsc{pop}}$ contains symbols corresponding to pops, while $\alphabet_{\textsc{push}}$ and $\alphabet_{\textsc{step}}$ contain symbols corresponding to pushes. $\alphabet_{\textsc{push}}$ and $\alphabet_{\textsc{step}}$ represent pushes in different ways: the former only includes the item that was pushed, whereas the latter additionally includes the premises and the rule that were used to trigger the push. 
Furthermore, define $\alphabet_1 \defeq \alphabet_{\textsc{pop}} \sqcup \alphabet_{\textsc{step}}$ and $\alphabet_2 \defeq \alphabet_{\textsc{pop}} \sqcup \alphabet_{\textsc{push}}$.
Note that $\alphabet_1$ and $\alphabet_2$ might be countably infinite if $\worldmodel$ is infinite.

We consider elements of $\kleene{\alphabet_1}$ and $\kleene{\alphabet_2}$, i.e., the Kleene closures over $\alphabet_1$ and $\alphabet_2$, respectively. Further note that the sets $\kleene{\alphabet_1}$ and $\kleene{\alphabet_2}$ form free monoids on $\alphabet_1$ and $\alphabet_2$, respectively.
Let $\lang_1 \subseteq \kleene{\alphabet}$ be the set of valid traces, i.e., traces that conform to the constraints of \cref{alg:A-star}, when representing pushes as proof steps. We use $\boldsymbol{\delta}=\delta_1 \circ \dots \circ \delta_{|\boldsymbol{\delta}|}$ to denote elements of $\lang_1$, where $\circ$ is the concatenation operator. 

We now explain how verbalizing execution traces in $\lang_1$ can be defined as a quotient set induced by a string homomorphism, contrasting \citeposs{lehnert2024beyond} approach to ours.
Let us first discuss \citet{lehnert2024beyond}. Their approach corresponds to the homomorphism $\homomorphl\colon \lang_1 \rightarrow \kleene{\alphabet_2}$ where $\homomorphl(\boldsymbol{\delta}) = \homomorphl(\delta_1) \circ \cdots \circ \homomorphl(\delta_{|\boldsymbol{\delta}|})$ and 
\begin{equation}
    \homomorphl(\delta) \defeq \begin{cases}
        \delta, \quad & \text{if}\ \delta \in \alphabet_{\textsc{pop}} \\
        \textsc{push}(\conclusion/\subst), \quad & \text{if}\ \delta = \textsc{push}((\premise_1/\subst, \dots, \premise_K/\subst), \arule, \conclusion/\subst) \in \alphabet_{\textsc{step}}.
    \end{cases}
\end{equation}
That is, $\homomorphl$ deletes the backpointers that store information about premises and rules from the trace.\footnote{We note that \citet{lehnert2024beyond} additionally include the values of $\weight$ and $\heuristic$, which we omit here for brevity.} 
Note that $\homomorphl$ is a homomorphism since it preserves the structure of concatenation, i.e., $\homomorphl(\boldsymbol{\delta}_1 \circ \boldsymbol{\delta}_2) = \homomorphl(\boldsymbol{\delta}_1) \circ \homomorphl(\boldsymbol{\delta}_2)$. The homomorphism induces a congruence $\sim_{\text{L}}$ as follows: $\boldsymbol{\delta}_1 \sim_{\text{L}} \boldsymbol{\delta}_2 \Leftrightarrow \homomorphl(\boldsymbol{\delta}_1) = \homomorphl(\boldsymbol{\delta}_2)$ \citep{howie1995fundamentals}. Formally, the set of what is verbalized is the quotient set $\lang_1/\sim_{\text{L}}$, which is the trace of the algorithm after deleting premises and rules. We remark that using this quotient would be inadequate for the purposes of our paper since it does not recover the proof. 

Now, the approach taken here corresponds to the following homomorphism: $\homomorph\colon \lang_1 \rightarrow \kleene{\alphabet_1}$, where $\homomorph(\boldsymbol{\delta}) = \homomorph(\delta_1) \circ \cdots \circ \homomorph(\delta_{|\boldsymbol{\delta}|})$ and
\begin{equation}
    \homomorph(\delta) \defeq \begin{cases}
        \varepsilon, \quad \text{if}\ \delta \in \alphabet_{\textsc{pop}} \\
        \delta, \quad \text{if}\ \delta \in \alphabet_{\textsc{step}}.
    \end{cases}
\end{equation}
That is, $\homomorph$ maps a trace sequence to the corresponding subsequence in which all pops have been deleted. With the congruence $\boldsymbol{\delta}_1 \sim \boldsymbol{\delta}_2 \Leftrightarrow \homomorph(\boldsymbol{\delta}_1) = \homomorph(\boldsymbol{\delta}_2)$, we verbalize the quotient set $\lang_1/\sim$,  which recovers the trace of the algorithm up to equivalence of pop orderings. We preserve the backpointers in order to be able to construct proofs and evaluate the correctness of the trace. From the backpointers, we can construct the minimal set of popped items required to yield the trace, as discussed in \cref{sec:verbalizing}.

\subsection{On the Choice of Cost Function}\label{sec:cost-function}
\paragraph{Superior Functions.} The cost function $\cost$ can be chosen freely as long as it is \defn{superior} \citep{knuth1977dijkstra}. That is, $\cost$ must have the following two properties: 
\begin{align}
    & \text{(i):}\quad \cost(\weight(\premise'_1), \dots, \weight(\premise'_k), \dots, \weight(\premise'_K)) \geq \cost(\weight(\premise'_1), \dots, \weight(\hat{\premise}_k), \dots, \weight(\premise'_K)\ \text{if}\ \weight(\premise'_k) \geq \weight(\hat{\premise}_k) \\
    & \!\text{(ii):}\quad \cost(\weight(\premise'_1), \dots, \weight(\premise'_K)) \geq \max_{k=1}^K\weight(\premise'_k).
\end{align}
We remark that a common alternative to \citeposs{knuth1977dijkstra} superior functions framework is to define weights using semirings; see, e.g., \citet{goodman-1999-semiring}, \citet{huang-2008-advanced}, and \citet{eisner-2023-time}. However, not all superior functions can be defined with semirings (and vice versa; see \citealp{gildea2021efficient}). Indeed, our particular choice of $\cost$ as $\cost(\weight(\premise'_1), \dots, \weight(\premise'_K)) \defeq 1+ \max_{k=1}^K\weight(\premise'_k)$, is one example. As mentioned in \cref{sec:a-star}, this yields a notion of shortest proof as the one with the shallowest depth. 
An alternative, reasonable choice is $\cost_2(\weight(\premise'_1), \dots, \weight(\premise'_K)) \defeq 1+ \sum_{k=1}^K\weight(\premise'_k)$, which yields a notion of shortest proof as the one with the fewest vertices---this cost function is well-defined under the semiring framework.
Note that both of these cost functions reduce to the shortest path for standard weighted graphs.\looseness=-1

\paragraph{Shallowest Proof vs. Proof with Fewest Vertices.} In this section, we compare the cost function
\begin{equation}\label{eq:cost-shallow}
    \cost_1(\weight(\premise'_1), \dots, \weight(\premise'_K)) \defeq 1+ \max_{k=1}^K\weight(\premise'_k)
\end{equation}
and the alternative
\begin{equation}\label{eq:cost-vertices}
    \cost_2(\weight(\premise'_1), \dots, \weight(\premise'_K)) \defeq 1+ \sum_{k=1}^K\weight(\premise'_k)
\end{equation}
in how they affect the execution traces returned by \cref{alg:A-star} and our verbalizations discussed in \cref{sec:verbalizing} and \cref{sec:equivalence-class}. Additionally, in order for \cref{eq:cost-vertices} to correspond to the number of vertices we require the weight of the axioms to be $1$, i.e., $\weight(\premise')=1$ for all $\premise'\in\axioms$, rather than $0$ as they are for \cref{eq:cost-shallow} (\cref{sec:a-star}). We measure verbalization length by summing the number of times an item has been verbalized in the trace over all pushes/proof steps. First note that with our notion of execution trace, i.e., with the homomorphism $\homomorph$ from \cref{sec:equivalence-class} over the alphabet $\alphabet_1 \defeq \alphabet_{\textsc{pop}} \sqcup \alphabet_{\textsc{step}}$ (see \cref{eq:push-vocab}), a pushed item $\conclusion'$ with $K$ premises will have a verbalization length of $2K + 2$: $K$ items as premises, $K+1$ items in the rule, and one conclusion.

First, consider a goal $\goal$ which has two proofs. One of them uses a single inference rule in one step over five axioms, and the other one uses a linear chain of four inference rules, each with a single premise and starting from a single axiom. The first proof is shortest according to $\cost_1$ (the first proof has cost $1$ and the second has cost $4$), whereas the second proof is the shortest according to $\cost_2$ (the first proof has cost $6$ and the second has cost $5$).
The first proof corresponds to the trace
\begin{equation}
    \trace_1 = \textsc{push}((\genitem{a}'_1, \dots, \genitem{a}'_5), \genitem{a}_1, \dots, \genitem{a}_5 \vdash \genitem{c}_{G}, \goal),
\end{equation}
which has a verbalization length of $12$ (we omit pushes of axioms since all axioms are pushed independently of the execution of \cref{alg:A-star}). The second proof corresponds to the trace
\begin{equation}
    \trace_2 = \textsc{push}((\genitem{b}'_1), \genitem{b}_1\vdash \genitem{b}_{2}, \genitem{b}'_{2}) \circ \cdots \circ \textsc{push}((\genitem{b}'_4), \genitem{b}_4\vdash \atom{c}_{G}, \goal),
\end{equation}
which has a verbalization length of $4\times 4 = 16$. So in this case, $\cost_1$ yields the proof that has the shortest verbalization length, which is desirable since it corresponds to a shorter token sequence generated by an LLM.

However, it is easy to see that this is not always the case. Consider a third proof, structured similarly to the second but with only two proof steps. This is now the new shortest proof according to $\cost_2$. It corresponds to the trace 
\begin{equation}
    \trace_3 = \textsc{push}((\genitem{b}'_1), \genitem{b}_1\vdash \genitem{b}_{2}, \genitem{b}'_{2}) \circ \textsc{push}((\genitem{b}'_2), \genitem{b}_2\vdash \genitem{c}_{G}, \goal),
\end{equation}
which has a verbalization length of $4\times 2 = 8$, i.e., shorter than that of the first proof.

We have demonstrated that neither $\cost_1$ nor $\cost_2$ dominates the other in terms of verbalization length and that the choice may make a difference in terms of which trace is returned by the algorithm. In our experiments, however, this would not make much difference, since the arity of the inference rules in ProofWriter and DeepRD is at most $2$. In fact, for the DeepRD data, the shortest proofs under the two cost functions always coincide.

\begin{figure}
    \centering
    \includegraphics[width=\linewidth]{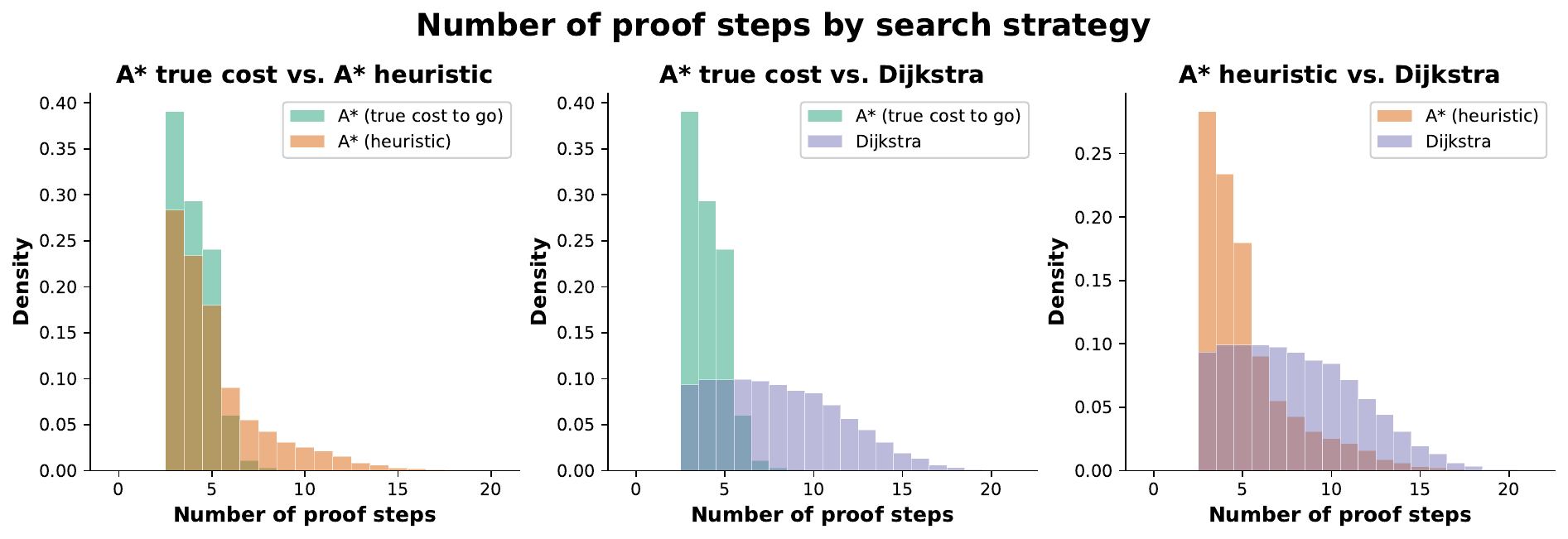}
    \vspace{-15pt}
    \caption{Comparison of number of proof steps between different search strategies. The histograms are computed based on the $9207$ problems in the ProofWriter training set. }
    \label{fig:proof_steps_comparison_proofwriter}
\end{figure}

\begin{figure}
    \centering
    \includegraphics[width=\linewidth]{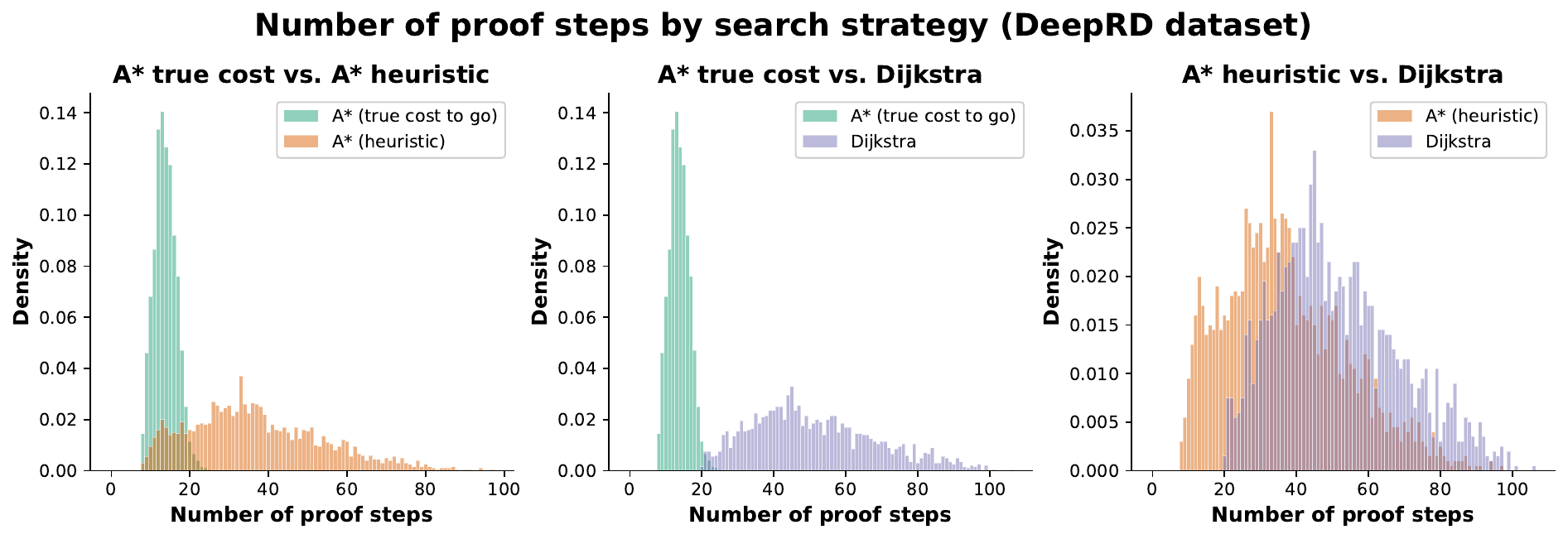}
    \vspace{-15pt}
    \caption{Comparison of number of proof steps between different search strategies. The histograms are computed based on the $2000$ problems in the DeepRD training set. }
    \label{fig:proof_steps_comparison_deeprd}
\end{figure}

\section{Experiment Details}\label{sec:details-experiments}

We give further details on the experiments and implementations.

\subsection{Efficiency Gains}\label{sec:efficiency-gains}
We provide an analysis on the number of proof steps (defined in \cref{sec:verbalizing}) taken by A* search with the dependency heuristic, A* search with the true cost-to-go, Dijkstra's algorithm for the two datasets. The results are shown in \cref{fig:proof_steps_comparison_proofwriter,fig:proof_steps_comparison_deeprd}. Our results on LLM post-training presented in \cref{sec:experiments} should be viewed in light of these analyses about the efficiency of the different algorithms on the two datasets.

\begin{figure}
    \centering
    \vspace{-6pt}
    \makebox[\textwidth][c]{%
        \hspace*{0\textwidth}%
        \includegraphics[width=1.05\textwidth]{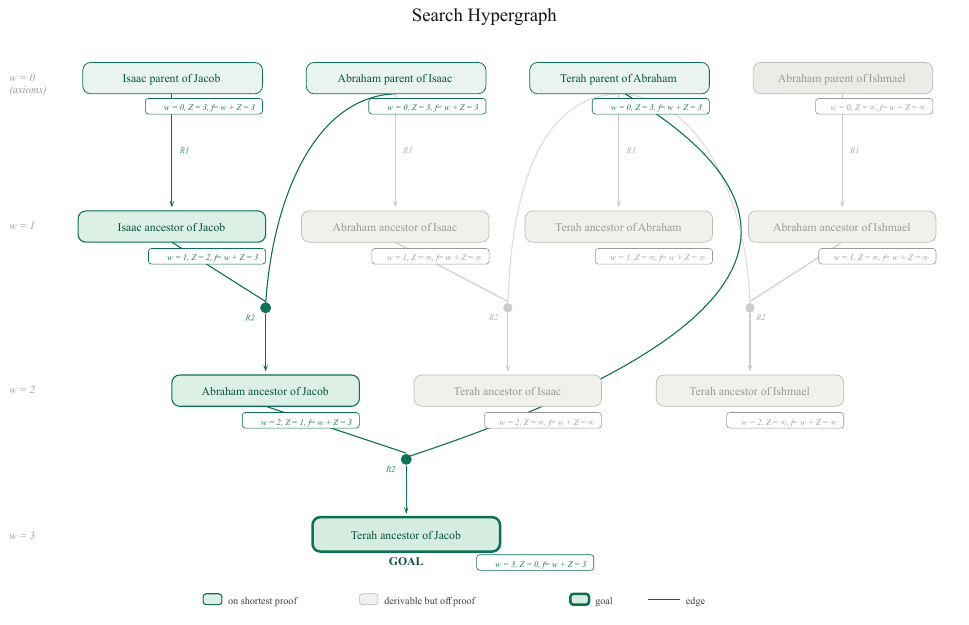}
    }
    \vspace{-14pt}
    \caption{Hypergraph induced by the logic program from \cref{fig:detailed-traces}, showing all 12 atoms in the minimal Herbrand model $\worldmodel$,      arranged by weight $\weight$ (i.e., depth). Each vertex is labeled with its  heuristic value $\heuristic$ under the true cost-to-go; on-path items (green) satisfy $\heuristic = \goalweight - \weight$, while off-path items receive $\heuristic = \infty$. The shortest proof of $\predicate{ancestor}(\const{terah}, \const{jacob})$ traces the unique path of finite $\heuristic$ values: $3 \to 2 \to 1 \to 0$.}
    \label{fig:hypergraph}
\end{figure}

\subsection{Prompt and Verbalizations}\label{sec:prompt-verbalizations}
We pass the following instruction prompt to the model:

\begin{promptbox}[title=Instruction Prompt]
You will be given a logic program verbalized in natural language, including a set of rules and a set of axioms. You will also be given a goal that is provable under this logic program. The formatting will be as follows:

Rules: <rule sentence 1>. <rule sentence 2>. ... <rule sentence n>.

Axioms: <axiom sentence 1>. <axiom sentence 2>. ... <axiom sentence m>.

Goal: Prove that <goal>.

Your task is to provide a correct proof of the goal. Each proof step must follow this format (one block per step) and be delimited by two new lines:\\

Premises: <premise sentence 1>. ... <premise sentence k>.

Rule: <rule sentence>.

Conclusion: <conclusion sentence>.\\

Terminate when the conclusion of a proof step is equal to the goal using the tags <answer></answer>. Here are a few examples of programs, goals, and proofs separated by --- for you to learn from:

\end{promptbox}

The last sentence above is only included for the ICL experiments. For those, we fix the (program, goal)-pairs across search orders and use $10$ ICL examples on ProofWriter and $5$ ICL examples on DeepRD (which are longer on average). The following is an example of a verbalized (program, goal)-pair with a verbalized proof from ProofWriter:

\begin{promptbox}[title=In-Context Example]
Rules: If X is blue, then X is furry. If X is nice, then X is furry. If X is blue and X is big, then X is nice. If X is cold, then X is quiet. If X is nice and X is furry, then X is cold. If gary is nice, then gary is smart. If X is cold, then X is furry. If X is cold and X is furry, then X is quiet.\\
Axioms: Bob is cold. Erin is nice. Gary is nice. Harry is blue.\\
Goal: Prove that gary is quiet.\\

Premises: Gary is nice.\\
Rule: If X is nice, then X is furry.\\
Conclusion: Gary is furry.\\

Premises: Gary is nice.\\
Rule: If gary is nice, then gary is smart.\\
Conclusion: Gary is smart.\\

Premises: Gary is nice. Gary is furry.\\
Rule: If X is nice and X is furry, then X is cold.\\
Conclusion: Gary is cold.\\

Premises: Gary is cold.\\
Rule: If X is cold, then X is quiet.\\
Conclusion: Gary is quiet.\\

<answer>Therefore, the goal is proven.</answer>
   
\end{promptbox}

\subsection{Training Details}\label{sec:training-details}

For the ProofWriter data we restrict the number of generated tokens to $1024$, whereas for the DeepRD data we restrict it to $2048$. These maximum lengths are long enough for the longest search traces under the A* dependency heuristic. 
DeepSeek-V3.2 is allowed up to $2048$ tokens for ProofWriter and $5000$ tokens for DeepRD.
All models are trained using the Adam optimizer \citep{2015-kingma-na-adam}. For SFT, we train for three epochs, storing checkpoints after each epoch and considering the one with the lowest loss on the validation set when evaluating on the test set. For SFT we use the learning rate $2\cdot 10^{-5}$ with $\beta_1 =0.9$ and $\beta_2=0.999$. We apply cosine learning rate scheduling with a warmup ratio of $0.03$. We use a batch size of $2$ for the 3B model and a batch size of $4$ for the 1B model. For GRPO, we train for one epoch with $16$ groups, KL temperature of $0.04$, learning rate of $1\cdot 10^{-6}$ with $\beta_1 =0.9$ and $\beta_2=0.999$. At inference, we generate solutions with greedy decoding.

\section{Additional Results}\label{sec:appendix-results}
\begin{figure}[t]
    \centering
    \includegraphics[width=0.99\linewidth]{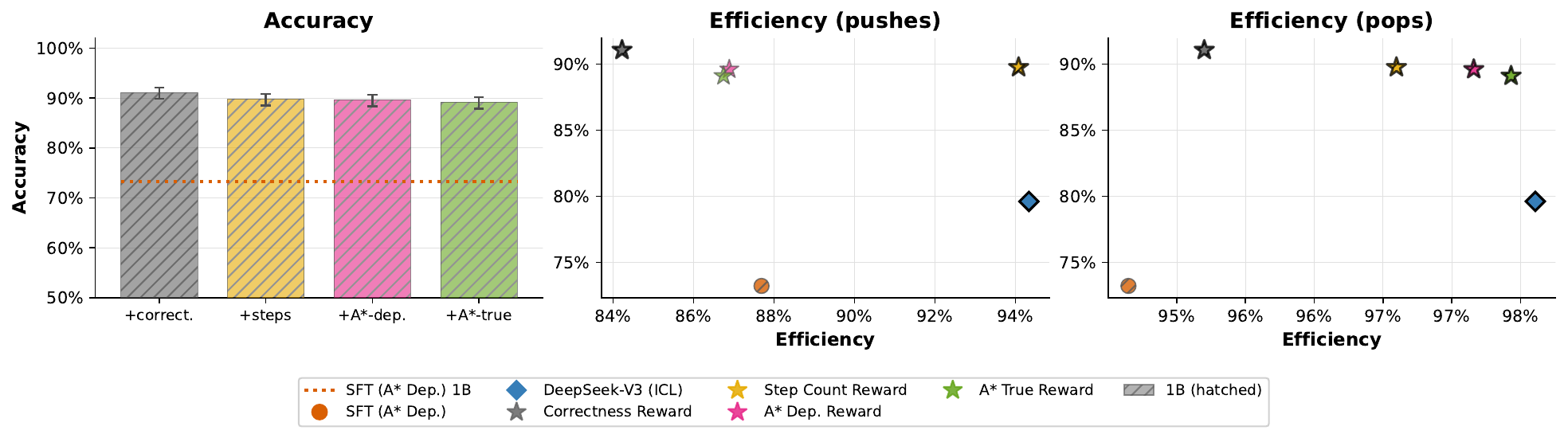}
    \caption{Results on \textbf{ProofWriter} for Llama-3.2-1B-Instruct finetuned with GRPO starting from the SFT checkpoint that was trained under the \emph{dependency heuristic}. (Note that \cref{fig:3b-grpo-results} shows results when considering the SFT checkpoint trainded under the true cost-to-go).}
    \label{fig:grpo-proofwriter-dependency}
\end{figure}

\begin{figure}[t]
    \centering
    \includegraphics[width=0.99\linewidth]{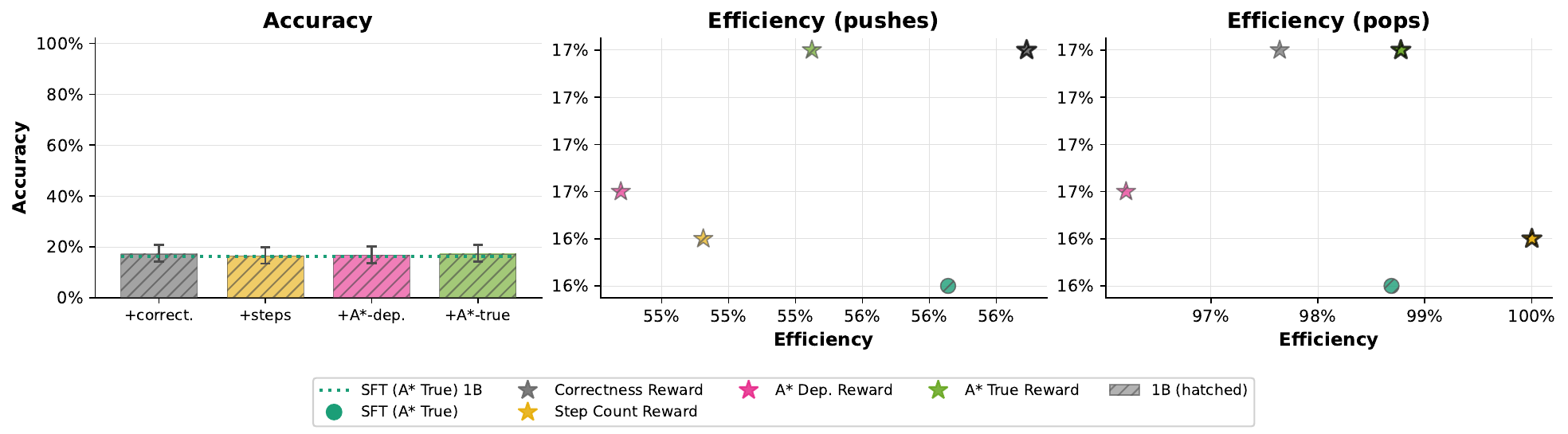}
    \caption{Results on \textbf{DeepRD} for Llama-3.2-1B-Instruct finetuned with GRPO starting from the SFT checkpoint that was trained under the \emph{true cost-to-go}. We note that RL is not effective in this case, in which the SFT reference model is weak on the task.} 
    \label{fig:grpo-deeprd-true}
\end{figure}

\cref{fig:grpo-proofwriter-dependency,fig:grpo-deeprd-true} show results for GRPO-finetuned Llama-3.2-1B-Instruct models. 
The results from \cref{fig:grpo-proofwriter-dependency} support our conclusions from \cref{fig:3b-grpo-results,fig:grpo-deeprd-dependency}---no reward model is overall dominating the others in terms of accuracy and efficiency. We also note that RL is unable to improve reasoning from the checkpoint trained on traces under the true cost-to-go on DeepRD (\cref{fig:grpo-deeprd-true}), which is a poor starting point; see \cref{fig:sft-results}. These results suggest  that RL is dependent on a good reference policy.

\end{document}